\documentclass{article}

\PassOptionsToPackage{numbers,compress}{natbib}

\usepackage[final,dblblindworkshop]{neurips_2026}

\usepackage[utf8]{inputenc}
\usepackage[T1]{fontenc}
\usepackage[hidelinks]{hyperref}
\usepackage{url}
\usepackage{xurl}
\usepackage{booktabs}
\usepackage{amsmath}
\usepackage{microtype}
\usepackage{graphicx}

\title{Preflight: Agentic Last-Mile Verification for AI Software Supply Chains}

\workshoptitle{AI for Science: Verification in the Age of AI Scientists}

\makeatletter
\renewcommand{\@noticestring}{Accepted at the AI for Science workshop (NeurIPS 2026).}
\makeatother

\author{%
\textbf{Yaman Jandali}\quad
\textbf{Nojan Sheybani}\quad
\textbf{Farinaz Koushanfar}\quad
\textbf{Tara Javidi}\\[3pt]
\normalfont\small Department of Electrical and Computer Engineering, UC San Diego\\
\normalfont\small La Jolla, California, USA
}

\begin{document}

\maketitle

\begin{abstract}
Software dependencies execute within the trust boundary of downstream
applications, yet provenance, static analysis, and package-level behavioral
approaches do not directly determine how a released artifact behaves for a
specific program. We present \textsc{Preflight}, a program-scoped pre-execution
verifier for Python dependencies that extracts application-referenced symbols,
iteratively synthesizes and validates function-level harnesses, executes resolved artifacts in a restricted environment, records behavioral fingerprints, and compares target
behavior against prior versions when suitable baselines exist. We evaluate
Preflight on four real software-supply-chain incidents under five application
contexts, a 100-package benign corpus, and a controlled function-level versus
import-only ablation. Preflight returns malicious verdicts in four incident
contexts and a clean verdict in the LiteLLM client context, where the affected
proxy-server path is not exercised. All 100 evaluated benign packages receive
clean verdicts, with zero observed false positives. In a controlled ablation on
Ultralytics and ctx, function-level harnesses expose malicious behavior in both
incidents, whereas import-only probing exposes neither. Together, these results
show that conditioning behavioral verification on the downstream program can
expose security-relevant dependency behavior while remaining selective on
benign workloads.
\end{abstract}

\section{Introduction}
\label{sec:introduction}

Python applications routinely execute third-party packages as black boxes within the
application's privilege context. This is particularly consequential for
AI-adjacent software, whose dependencies may have access to model-provider
tokens, cloud credentials, local files, development environments, and user
data. A compromised dependency can therefore execute inside an application's
existing trust boundary.

Package identity, provenance, and source inspection provide important security
signals, but they do not directly establish how a released artifact behaves in
the context of the particular program about to execute it. Relevant behavior
may occur during installation or import, or remain dormant until a specific
function, constructor, plugin path, argument pattern, or environment condition
is exercised. Conversely, an application may use only a small portion of a
package's API surface. We refer to this gap between package-level assurance and
application-specific execution as the \emph{last mile} of software-supply-chain
verification.

We therefore formulate dependency verification as a program-scoped problem.
For program $A$ with pinned dependency state $L$, the verifier evaluates whether
version $v$ of package $P$ exhibits security-sensitive behavior through the
application-referenced dependency surface extracted from $A$, and, when a
suitable baseline exists, whether that behavior is newly introduced relative to
the same application surface in a prior version. This formulation binds the
decision to the program, lockfile, artifact, referenced API surface, and
intended execution.

We present \textsc{Preflight}, a pre-execution verifier that extracts imported
packages and referenced package symbols, synthesizes harnesses that exercise
those symbols, and executes resolved artifacts in a restricted Docker sandbox.
Preflight records a canonical behavioral fingerprint containing network
attempts, sensitive environment access, filesystem writes, subprocess creation,
and relevant syscall classes. It then computes capability deltas against exact
prior versions where available. Static findings provide additional evidence,
while signed certificates bind the final decision to the analyzed program and
dependency state.

Large language models are used in a bounded agentic role. Harness construction is a closed loop in which the model proposes a probe, a deterministic validator accepts or rejects it, and rejected probes are revised under validator feedback within a fixed retry budget before the pipeline falls back to a deterministic harness. During harness synthesis, the model's action space is limited to emitting and revising harness source. It does not select packages, choose sandbox policy, interpret raw traces, or decide when analysis terminates. Application-surface extraction, harness validation, sandbox orchestration, trace collection, fingerprint construction, version differencing, deterministic fallback logic, certificate signing, and certificate verification remain deterministic.

This paper makes three contributions. First, we introduce a program-scoped
dependency-verification architecture combining downstream program context,
sandboxed behavioral observation, prior-version differencing, and signed
pre-execution certificates. Second, we use a bounded agentic synthesis loop, in which model-generated harnesses are revised under deterministic validator feedback, to exercise application-relevant dependency functionality while retaining deterministic validation and enforcement. Third, we evaluate Preflight on four real
software-supply-chain incidents under five application contexts, a 100-package
benign corpus, and a controlled function-level versus import-only ablation.
Across these application contexts, Preflight reports four malicious verdicts
and a clean verdict for the LiteLLM client workload, which does not exercise
the affected proxy-server path. The benign corpus yields
zero false positives, and the ablation shows that function-level probing exposes
malicious behavior in both evaluated cases where import-only execution does not.

\section{Related Work}
\label{sec:related-work}

\paragraph{Provenance and attestation.}
Provenance systems bind software artifacts to build steps, identities, and
attestations. in-toto provides a general supply-chain attestation framework,
and SLSA specifies supply-chain guarantees and provenance formats
\cite{torresarias2019intoto,slsa_spec_v12}. PyPI supports digital attestations
including PyPI Publish and SLSA Provenance predicates
\cite{pypi_attestations_intro,pypi_attestations_security}. These mechanisms
establish properties of artifact origin and production. Preflight complements
them with application-conditioned behavioral evidence.

\paragraph{Static and dynamic package analysis.}
Bandit, Semgrep, and GuardDog provide static or metadata-oriented security
analysis \cite{bandit_docs,semgrep_repo,guarddog_github}, while MALOSS and
Amalfi study malicious-package detection at ecosystem scale
\cite{duan2021maloss,sejfia2022amalfi}. Dynamic systems instead execute packages
and inspect runtime behavior. OpenSSF Package Analysis provides registry-scale
behavioral analysis. DONAPI and DySec use dynamic behavior in malicious-package
detection \cite{openssf_package_analysis,huang2024donapi,mehedi2026dysec}, and
Packj combines static vetting with dynamic tracing and sandboxing
\cite{packj_tool}. Preflight differs primarily by conditioning dynamic
execution on a particular downstream program's dependency surface.

\paragraph{Differencing and reachability.}
Prior work applies differential static analysis to suspicious package updates
and studies runtime restrictions for malicious dependencies
\cite{froh2023differential,ferreira2021permissions}. Mininode and DepClean
identify unused dependencies or code to reduce attack surface
\cite{koishybayev2020mininode,sotovalero2020depclean}, while Semgrep Supply
Chain and FIKA use reachability or executable evidence to refine dependency
findings \cite{semgrep_supply_chain_reachability,gamage2026fika}. Preflight
combines application-derived dependency references with behavioral version
differencing over released artifacts.

\paragraph{LLM-assisted test and harness generation.}
TestPilot, ASTER, MuTAP, and CoverUp use language models to generate tests or
increase coverage using program context, static analysis, mutation feedback, or
coverage feedback
\cite{schaefer2023testpilot,pan2025aster,dakhel2024mutap,pizzorno2025coverup}.
Preflight applies related synthesis techniques to dependency verification:
model-generated probes exercise application-referenced APIs, while deterministic
validation, behavioral tracing, and certificate enforcement surround the
synthesis step. These systems close their loops over coverage or mutation signals from the program under test. Preflight closes over validator acceptance of application-conditioned probes against an untrusted artifact.

\paragraph{Malicious-package corpora.}
Reproducible supply-chain evaluation relies on identifiable malicious artifacts
and incident labels. Relevant resources include Backstabber's Knife Collection,
Datadog's human-vetted malicious-package dataset, the OpenSSF
malicious-packages repository and OSV's MAL namespace, and execution-oriented
datasets such as OSPtrack and QUT-DV25
\cite{ohm2020backstabbers,datadog_malicious_dataset,
openssf_malicious_packages,osv_malicious_api,osptrack2025,qutdv25_2025}.
Incident reports for LiteLLM and Ultralytics provide additional documentation
for compromised releases used in our evaluation
\cite{litellm_issue_24518,ultralytics_issue_18027}.

\section{Methodology}
\label{sec:methodology}

\begin{figure}[t]
\centering
\includegraphics[width=\linewidth]{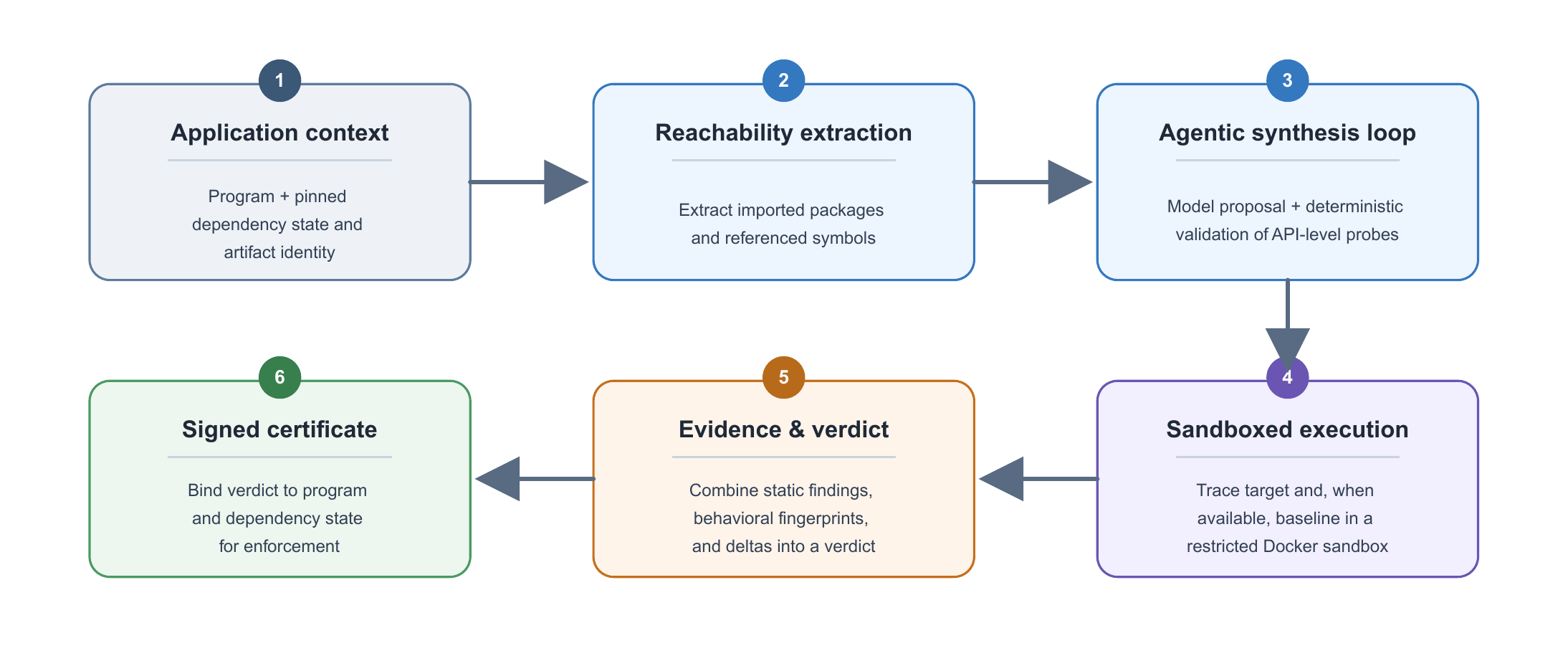}
\caption{Overview of the Preflight verification pipeline. Given a downstream
program and pinned dependency state, Preflight synthesizes validated
application-conditioned probes, observes target and baseline behavior when a
suitable baseline exists, combines behavioral, differential, and static
evidence, and emits a signed certificate for enforcement.}
\label{fig:preflight-overview}
\end{figure}

Figure~\ref{fig:preflight-overview} summarizes the Preflight pipeline.

\subsection{Threat Model and Verification Goal}
\label{sec:threat-model}

Preflight treats the downstream program and its pinned dependency specification
as the intended execution state and resolved dependency artifacts as untrusted.
An adversary may compromise a legitimate package release or publish a malicious
update whose behavior executes during installation, import, or subsequent API
invocation.

Let $A$ denote the downstream application, $L$ its pinned dependency state, and
$P_v$ version $v$ of dependency $P$. Preflight evaluates behavior of $P_v$
through the application-referenced dependency surface extracted from $A$. When
a suitable reference version $P_b$ is available, the primary differential
signal is whether $P_v$ introduces security-sensitive capabilities absent when
the same application-facing functionality is exercised against $P_b$.

The verdict is therefore program-scoped rather than package-global. The same
package version can yield different decisions for applications that exercise
different portions of its API. Preflight targets malicious or otherwise
security-relevant dependency behavior rather than conventional application
vulnerabilities, and its certificate records the program and dependency state
for which the analysis was performed.

\subsection{Application-Surface Extraction and Harness Generation}
\label{sec:reachability}

Preflight parses the downstream Python program using Python's abstract syntax
tree (AST) and identifies imported packages together with package symbols
referenced by the program. The analysis handles direct imports, from-imports,
aliases, module-attribute references, and cases in which a package is imported
without an explicit function-level reference. These records are joined with
pinned versions from requirements.txt, uv.lock, or Pipfile.lock. Unpinned
dependencies are rejected because verification must bind to a specific package
version.

The extracted references define the application-referenced dependency surface
that Preflight attempts to exercise. For example, if an application invokes
requests.get, the corresponding probe attempts to invoke that API rather than
merely importing requests. This focuses dynamic analysis on referenced
constructors, functions, and other runtime entry points instead of relying only
on package import.

For each application-referenced dependency surface, Preflight synthesizes a
small executable harness using OpenAI's Responses API and
\texttt{gpt-5.4-mini} (temperature 0, maximum 4,096 output tokens, 120-second
API timeout). The model receives package-source context, package identity and
version, referenced symbols, and invocation constraints, then proposes Python
code to exercise those symbols. The prompt templates are retained in the
implementation. The model alias was recorded without an immutable snapshot or
seed; other sampling parameters were not explicitly set.

Each candidate undergoes at most two validator rounds and must be valid Python
of at most 100 lines, import the target package, invoke all referenced symbols,
respect fixed invocation constraints, and preserve symbol coverage after AST
instrumentation. Rejection feedback guides one revision; runtime traces are
not fed back to the model. API calls permit up to three attempts on transient
errors. Ordinary operation permits deterministic harness fallback, but the
live-agent evaluation excludes fallback generations. Across the five reported
application contexts, all nine final harnesses were validated and executed:
four passed on the first candidate, and five required one revision. This used
14 API calls (18,631 input and 2,890 output tokens), with approximately
\$0.027 in estimated generation cost and no deterministic harness fallbacks.

Preflight also performs pre-execution analysis using Bandit, Semgrep, a
typosquat checker, and supply-chain-specific Semgrep rules. Static findings are
incorporated into the verdict alongside behavioral and version-differential
evidence. The core program-scoped analysis remains behavioral, although certain
critical static findings can independently produce a malicious verdict.

\subsection{Sandboxed Behavioral Observation}
\label{sec:sandbox}

Preflight installs and executes each artifact inside a restricted Docker
environment. Package code runs as a non-root user with a read-only root
filesystem, no-new-privileges enabled, dropped Linux capabilities, bounded
resources, and a seccomp policy. Because tracing relies on strace, the sandbox
permits the ptrace functionality required for instrumentation. Both package
installation and harness execution are traced. Preflight records process and
system-call activity with strace, captures network traffic, logs DNS activity
through an internal sinkhole, and compares filesystem state before and after
execution.

Raw traces contain temporary paths, installer activity, process identifiers,
and other execution-specific noise. Preflight reduces these observations to a
canonical behavioral fingerprint containing normalized endpoints, sensitive
environment reads, persistent writes, spawned executables, and syscall
categories. Raw evidence remains available for inspection, while the canonical
representation supports stable version-to-version comparison.

\subsection{Differential Verification}
\label{sec:differential}

Network access, filesystem writes, and subprocess creation are common in benign
software, so Preflight treats behavioral change across versions as a primary
signal rather than classifying these capabilities in isolation. Let $F(P,A)$
denote the set of canonical capabilities observed when package $P$ is exercised
through the application-referenced dependency surface extracted from $A$. For
target version $P_v$ and baseline $P_b$, Preflight computes

\[
\Delta_A(P_v,P_b) = F(P_v,A) \setminus F(P_b,A).
\]

The differential includes newly observed endpoints, sensitive environment
reads, persistent writes, subprocess executables, syscall classes, and changes
to the exercised API surface. The policy assigns higher severity to selected
changes such as credential-bearing environment access, previously unseen or
undeclared endpoints, persistent writes, and newly observed subprocesses.

When a baseline version is explicitly specified, Preflight uses that exact
version and records the target and baseline artifact identities. When no
defensible predecessor is available, Preflight performs target-only analysis
and reports the observed behavioral and static evidence. Claims about newly
introduced behavior are reserved for controlled target--baseline comparisons.

\subsection{Verdicts and Certificates}
\label{sec:verdicts}

Preflight produces \emph{clean}, \emph{suspicious}, or \emph{malicious}
verdicts from the target fingerprint, differential evidence, application-surface
information, and static findings. Optional model-assisted verdict synthesis
using \texttt{gpt-5.4} was attempted for each of the five incident contexts,
but every request exceeded the API input-size or context limit. Consequently,
all five final incident verdicts used deterministic fallback, not LLM verdict
judgments. Under this policy, at least two critical capability findings imply
\emph{malicious}, one implies \emph{suspicious}, and none implies \emph{clean},
unless a critical static finding overrides the result. High-severity static
findings alone are not critical. In target-only analysis, observed capabilities
are compared with an empty reference set rather than a benign predecessor.

The result is encoded as an in-toto statement in a Dead Simple Signing Envelope
(DSSE) and signed with Ed25519. A separate checker validates the signature,
statement and subject digests, certificate freshness, program hash, lockfile
hash, and verdict, and permits execution only when the certificate is valid and
carries a clean verdict. This separates evidence generation from a small
deterministic enforcement step.

\section{Evaluation}
\label{sec:evaluation}

\subsection{Experimental Setup}
\label{sec:eval-setup}

We evaluate Preflight along three dimensions: program-scoped behavior on real
compromised-package incidents, false positives on benign packages, and
function-level versus import-only probing. The incident analyses report the
behavioral evidence underlying each verdict, while the ablation isolates
harness mode.

For the incident cases, we separately track a strict expected-findings oracle,
which is satisfied only when every preregistered behavioral category is
observed. This measures behavioral coverage independently of the detector
verdict.

The incident evaluation uses four real supply-chain incidents (LiteLLM,
Ultralytics, ctx, and DurableTask) under five application contexts. Incident
artifacts are obtained from provenance-controlled sources, including retained
PyPI artifacts and a pinned snapshot of the Datadog malicious-package dataset
\cite{datadog_malicious_dataset}. We evaluate an incident artifact only when its
exact package identity and artifact hash can be established. Exact version
differentials use LiteLLM $1.82.6 \rightarrow 1.82.7$, Ultralytics
$8.3.40 \rightarrow 8.3.41$, and ctx
$0.1.2 \rightarrow 0.2.2$. DurableTask is evaluated target-only because no
defensible benign predecessor is used. Controlled differentials use
hash-identified artifacts with matched dependency treatment across target and
baseline.

LiteLLM contributes two application contexts over the same package pair. The
preregistered client workload exercises ordinary model-client functionality and
does not enter the compromised proxy-server path. We additionally evaluate the
same versions with a separately labeled exploratory proxy-server workload that
exercises the affected functionality. Together, these workloads compare two
application surfaces while holding the target and baseline versions fixed.

We report only provenance-complete, evaluable incident cases. A non-evaluable
SentinelOne attempt and torchtriton, for which the exact malicious artifact
could not be established, are excluded.

For the benign evaluation, we use 100 pinned AI/data/ML-adjacent packages
curated from PyPI and measure the fraction receiving a suspicious or malicious
verdict. All 100 executions complete with clean verdicts and no infrastructure
errors.

For the function-level versus import-only ablation, Ultralytics and ctx are each
executed with a model-generated function-level harness and an import-only
harness. Target and baseline artifacts, dependency closures, referenced-symbol
specifications, environment, containment, timeouts, detector policy, and
evaluation code are held fixed - only harness mode changes.

\subsection{Program-Scoped Incident Case Studies}
\label{sec:malicious-cases}

Table~\ref{tab:malicious} summarizes the five application contexts. Preflight
returns four malicious verdicts and one clean verdict. The LiteLLM client
workload does not exercise the compromised proxy-server path, while the
separately labeled exploratory proxy workload does.

\begin{table}[t]
\caption{Program-scoped verdicts across four real supply-chain incidents and
five application contexts. LiteLLM uses the same package pair under two
workloads: only the proxy workload reaches the compromised path. DurableTask is
evaluated target-only because no defensible benign predecessor is used.}
\label{tab:malicious}
\centering
\begin{tabular}{lll}
\toprule
Context & Comparison / scope & Verdict \\
\midrule
LiteLLM client & $1.82.6 \rightarrow 1.82.7$ & clean \\
LiteLLM proxy & $1.82.6 \rightarrow 1.82.7$ & malicious \\
Ultralytics & $8.3.40 \rightarrow 8.3.41$ & malicious \\
ctx & $0.1.2 \rightarrow 0.2.2$ & malicious \\
DurableTask & target-only & malicious \\
\bottomrule
\end{tabular}
\end{table}

\paragraph{Verdict attribution.}
Static findings did not drive any of these five verdicts. Critical capability
counts were zero for LiteLLM client; 14 for LiteLLM proxy (two metadata
endpoints and 12 executables); six for Ultralytics (three endpoints and three
\texttt{curl} paths); and three for ctx (endpoints). DurableTask's target-only
verdict arose from three endpoints: \texttt{check.git-service.com}, the sandbox
sinkhole, and Docker DNS. Two counted endpoints were sandbox infrastructure;
without a benign predecessor, the verdict cannot establish a malicious
version-to-version change.

\begin{figure}[!ht]
\centering
\includegraphics[width=\linewidth]{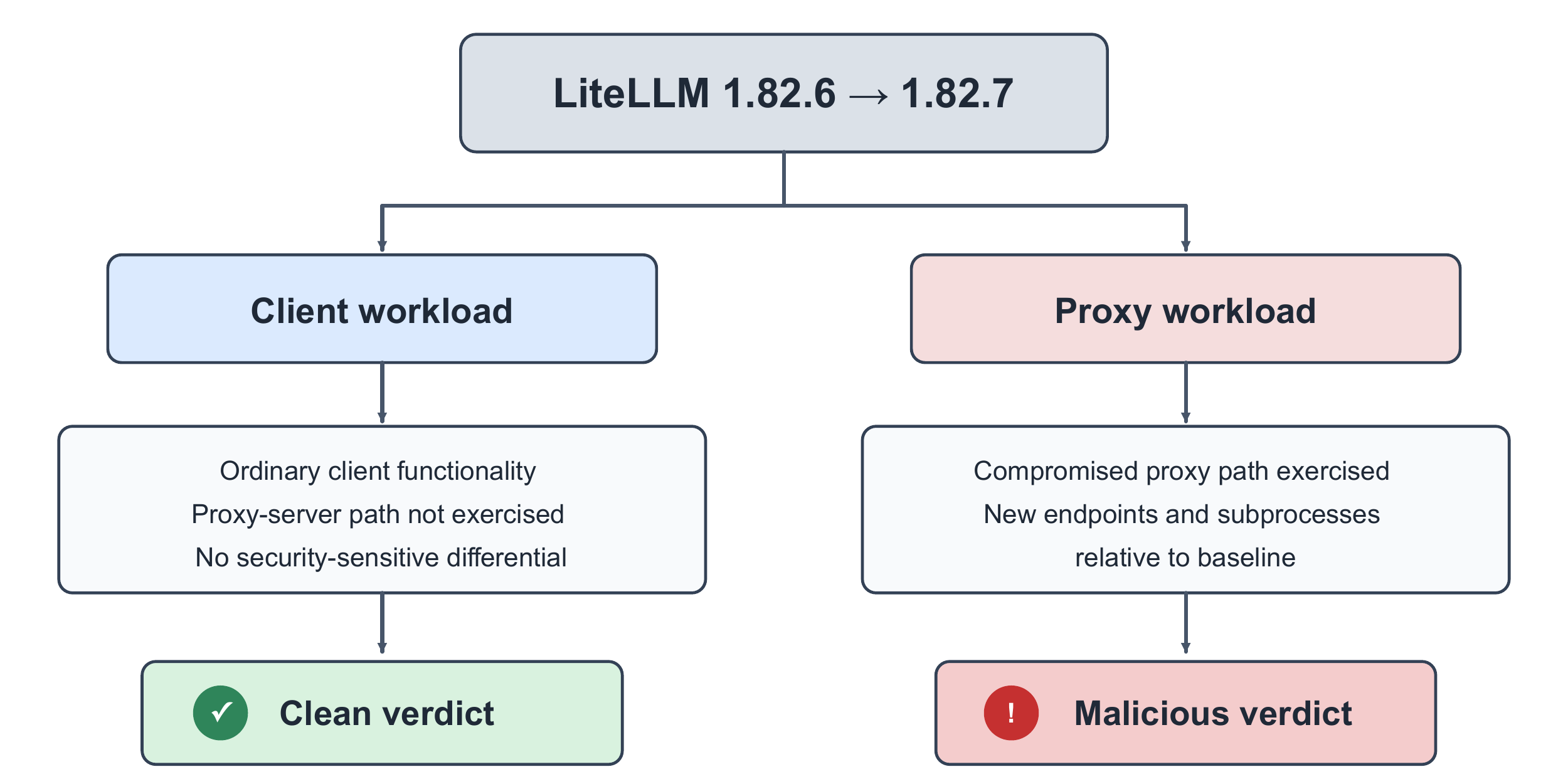}
\caption{LiteLLM under two application contexts with the package pair fixed.
The client workload does not reach the compromised proxy path and receives a
clean verdict. The proxy workload reaches it, exposes new endpoints and
subprocesses, and receives a malicious verdict.}
\label{fig:litellm}
\end{figure}

\paragraph{LiteLLM.}
The two LiteLLM workloads isolate application scope while holding the package
pair fixed. The client workload does not enter the compromised proxy-server
path and produces no security-sensitive differential, yielding a clean verdict.
The proxy workload reaches the affected path, exposing network endpoints and
subprocesses in the target that are absent from the baseline and yielding a
malicious verdict. The expected endpoint finding is observed, while the
expected environment-read and persistent-write findings are not. Expected-finding
coverage is tracked separately from the detector verdict.
Figure~\ref{fig:litellm} illustrates this contrast.

\paragraph{Ultralytics.}
The Ultralytics workload exercises the application-referenced YOLO model
constructor, the package's model-instantiation interface. Relative to version
8.3.40, compromised version 8.3.41 introduces a new endpoint and subprocess,
producing a malicious verdict with both preregistered findings observed.
Import-only execution exposes neither finding.

\paragraph{ctx.}
For ctx, version 0.1.2 serves as the benign predecessor for compromised version
0.2.2. The generated harness invokes the application-referenced Ctx constructor.
Preflight observes a network endpoint in the target version that is absent from
the baseline and returns malicious. The expected endpoint finding is observed -
the expected environment-read finding is not.

\paragraph{DurableTask.}
DurableTask's source contains a downloader, but execution showed only domain
resolution and sinkholed connections, not the expected persistent write or
subprocess. Neither preregistered category was matched, and completed payload
execution was not established.

\subsection{Benign-Package Evaluation}
\label{sec:benign}

We evaluate Preflight on 100 pinned AI/data/ML-adjacent packages curated from
PyPI. All 100 receive clean verdicts with zero false positives and zero
infrastructure errors, for an observed false-positive rate of 0/100 (0\%).
This shows that Preflight remains selective on the evaluated benign workloads
despite monitoring broad classes of security-relevant behavior. The measurement
applies to the exercised workloads rather than all possible API paths.

\subsection{Function-Level vs.\ Import-Only Ablation}
\label{sec:ablation}

The ablation compares model-generated function-level harnesses with import-only
probing on two incident cases, Ultralytics and ctx, while holding the remaining
experimental inputs fixed.

\begin{table}[t]
\caption{Function-level versus import-only ablation on two incident cases. All
experimental inputs except harness mode are held fixed.}
\label{tab:ablation}
\centering
\begin{tabular}{lcc}
\toprule
Case & Function-level harness & Import-only probe \\
\midrule
Ultralytics & malicious & clean \\
ctx & malicious & clean \\
\midrule
Malicious verdicts in ablation & 2/2 & 0/2 \\
\bottomrule
\end{tabular}
\end{table}

Function-level harnesses produce malicious verdicts for both incidents, whereas
import-only probes produce clean verdicts for both. Function-level execution
observes both preregistered findings for Ultralytics and one of two for ctx. Import-only probing observes none of the preregistered findings in either case.

This provides direct mechanistic evidence that exercising
application-referenced functionality can expose malicious behavior missed by
import-only execution. The comparison isolates function-level versus import-only execution. It does not isolate iterative synthesis from single-shot generation, nor model-generated harnesses from deterministic ones supplied with equivalent API knowledge.

\section{Discussion and Limitations}
\label{sec:discussion}

Taken together, the evaluation provides mutually reinforcing evidence for
Preflight's central design: program-scoped verification tracks behavior exposed
through the exercised application surface, remains selective on the evaluated
benign corpus, and function-level probing exposes behavior that import-only
execution misses in both ablation cases.

\paragraph{Program-scoped semantics and coverage.}
LiteLLM provides the clearest controlled example of these semantics. Holding
the package pair fixed, the client workload receives a clean verdict because it
does not enter the compromised proxy path, whereas the proxy workload reaches
that functionality and receives a malicious verdict. The verification decision
therefore reflects the application surface being exercised rather than package
identity alone.

Reflection, dynamically constructed imports, callbacks, plugins,
input-sensitive branches, and environment-dependent behavior can fall outside
the extracted or exercised surface. Richer program analysis and finer-grained
harness synthesis could extend the exercised application surface.

\paragraph{Verdicts, oracle coverage, and containment.}
LiteLLM proxy and ctx produce malicious verdicts while matching subsets of
their preregistered expected findings, and DurableTask produces a malicious
verdict without matching either expected category. Preflight therefore records
detector verdict and expected-finding coverage independently, distinguishing
the security decision from completeness of preregistered telemetry coverage.

Containment can prevent later stages of malicious execution after earlier
behavior has already been observed. Raw traces remain important for interpreting
such cases and for distinguishing behavior that appears through a different
telemetry category. The current canonical representation does not provide
complete phase attribution among installation, import, and explicit API
execution.

\paragraph{Baselines and provenance.}
A prior version is a behavioral reference rather than proof of benignity, so
baseline selection and exact artifact identity remain important. Target-only
cases such as DurableTask support claims about observed behavior but not about
capabilities newly introduced by a particular version.

\paragraph{Scaling the evaluation.}
The current evaluation establishes the central program-scoped mechanism across
multiple real incidents and a larger complementary benign corpus. A natural
next step is to scale the incident evaluation across additional attack classes,
package architectures, and downstream applications.

The current harness-generation stage targets application-referenced symbols at
the function or constructor level. Finer-grained harness synthesis could
generate multiple probes per referenced symbol, explore different argument
configurations, object states, and call sequences, and use execution feedback
to expand behavioral coverage. This would preserve Preflight's program-scoped
semantics while exercising a richer set of behaviors within the
application-referenced dependency surface.

\paragraph{Model role and isolation.}
The agentic component is scoped to harness construction: the loop closes over deterministic validator feedback, not over observed runtime behavior. Feeding execution traces back into probe selection would extend the loop to behavioral coverage and remains future work. The ablation isolates function-level versus import-only execution. It does not
establish an advantage of LLM-generated harnesses over deterministic
function-level harnesses supplied with equivalent API knowledge. The model
instead serves to construct application-conditioned runtime stimuli, while
deterministic components validate harnesses, collect evidence, enforce policy,
and verify certificates.

Preflight currently executes untrusted artifacts inside a restricted Docker
environment. Stronger isolation, such as disposable microVMs, can be
incorporated without changing the program-scoped verification architecture.

\section{Conclusion}
\label{sec:conclusion}

We presented \textsc{Preflight}, a program-scoped pre-execution verifier that
combines application-referenced dependency analysis, function-level runtime
probing, sandboxed behavioral observation, version differencing, and signed
verification certificates. Across four real software-supply-chain incidents
and five application contexts, Preflight reports four malicious verdicts and a
clean verdict for the LiteLLM client workload, which does not exercise the
compromised proxy-server path. The 100-package benign
evaluation yields zero false positives, while the controlled Ultralytics and ctx
ablation shows that function-level harnesses expose malicious behavior in both
cases where import-only probing exposes neither.

These results support program-scoped behavioral verification as a practical
last-mile layer between package-level assurance and the program about to execute
a dependency. Broader incident evaluation and finer-grained harness synthesis
provide natural directions for extending the approach to additional attack
classes and more complex application behaviors.

\bibliographystyle{plainnat}
\bibliography{ref}

\end{document}